\pdfoutput=1
\documentclass[11pt,a4paper]{article}
\usepackage[margin=1in]{geometry}

\usepackage[]{acl}
\usepackage{amssymb}

\usepackage{times}
\usepackage{latexsym}

\usepackage[T1]{fontenc}
\usepackage[utf8]{inputenc}

\usepackage{microtype}
\usepackage{graphicx}

\providecommand{\wgs}[1]
{\noindent
  \begin{small}	
  \textbf{\textit{Relevant UniDive working groups:~}} #1
  \end{small}
}

\usepackage{gb4e}
\noautomath
\title{The \textit{KIPARLA Forest} treebank of spoken Italian: an overview of initial design choices}

\author{Ludovica Pannitto \\
  LILEC - University of Bologna \\
  \texttt{ludovica.pannitto@unibo.it} \\\And
  Caterina Mauri \\
  LILEC - University of Bologna\\
  \texttt{caterina.mauri@unibo.it} \\}

\begin{document}
\maketitle

\wgs{WG1, WG4}

\section{Introduction}

The KIParla corpus\footnote{\url{www.kiparka.it}} \cite{maurikiparla,ballare2020creazione} is an existing and well-known resource for spoken Italian: the project stems from the imperative to capture the linguistic diversity inherent in spoken language, representing variation with respect to the properties of both the interaction and the speakers involved.
The corpus encompasses a diverse range of Italian spoken varieties, manually transcribed following Jefferson guidelines\footnote{\url{https://benjamins.com/catalog/pbns.125.02jef}} and aligned with audio files: the resource is structured in an incremental and modular fashion, which allows for the addition of new corpus modules over time, focusing on different dimensions of linguistic variation and geographical areas. It is already freely available for consultation, and an orthographic transcription is provided upon request.
Each module within the overarching KIParla project is balanced to ensure its coherence and self-sufficiency as a linguistic resource. All summed up, the KIParla counts ca. 228 hours of recordings and approximately 1,990,311 transcribed tokens.

Generally, spoken data are rare in the broader context of Universal Dependency (henceforth, UD) treebanks \cite{Dobrovoljc2022} but offer unique insights into linguistic phenomena and support various research fields, ranging from typology to NLP. More specifically, among existing Italian UD treebanks, none is explicitly addressing spoken varieties and those that contain some spoken language \cite{delmonte2007vit,sanguinetti2015parttut,agnoloni2022making} do not focus on variation. We are working on aligning the existing KIParla annotation to UD format and start building morphosyntactic annotation on top of the existing resource.
Preliminary linguistic annotation efforts on the KIParla corpus (see EVALITA\footnote{\url{https://www.evalita.it/}} evaluation campaign in 2020, \citealt{Bosco2020}) highlighted the challenges in tagging unconstrained speech contexts. %The newly collected modules, characterized by regional variation and code-switching phenomena, are bound to introduce further challenges.
The aim of this preliminary work is to examine choices taken for existing spoken UD resources and try to align our resource to the existing ones. In the remainder of this abstract, we will follow the points raised in \citealt{Dobrovoljc2022} in order to touch all relevant aspects.
At the moment, conversations are manually transcribed by student interns through the \textsc{elan}\footnote{\url{https://archive.mpi.nl/tla/elan}} software in \texttt{.eaf} format. 
Transcribers also segment the transcription into basic intonational units, which we refer to as \textit{transcription units} (henceforth, TUs). Trascriptions are then revised and anonymized by an expert. The first step towards the construction of the treebank consists therefore in fully transforming the current Jeffersonian notation into \texttt{conll} and later \texttt{conll-u} notation.
%At the end of the current process, each TU looks like this:
%\begin{exe}
%\ex 
%\gll ((ride)) (i)nfatt[i: >c]io[è< s'è pagato me:no]\\
%%cat.NOM eat.3.SG.PRS sour-cream.ACC sour-cream.ACC sour-cream.ACC sour-cream.ACC\\
%\trans `((laughs)) actually I mean we paid less'
%\end{exe}

\section{Points of interest}

\subsection{Speech Specific Metadata}

Most spoken treebanks include speech-specific metadata such as link to soundfile, information about the speaker and information on language variety. We provide \textbf{link to sound file} at the level of TU metadata (\texttt{\# sound\_url}): this will point to a permanent link which is however only accessible to registered users, due to privacy reasons. Two attributes (\texttt{AlignBegin} and \texttt{AlignEnd}, expressed in milliseconds) are provided at token level in the \texttt{MISC} field. Typically, these will be valorized on the first and last token of each TU. Similarly, each TU will include the \textbf{speaker ID} as metadata (\texttt{\# speaker\_id}). Each speaker is then described through its metadata (including data such as gender, age, origin, education level, etc.) in a separate \texttt{json} file. The same is true for conversation-specific metadata (i.e., number of participants, place and date of recording, type of interaction event...).

As far as \textbf{language variety} is concerned, Frisian-Dutch Fame \cite{Braggaar2021} and Turkish-German SAGT \cite{ccetinouglu2019challenges} use the \texttt{Lang} attribute in the \texttt{MISC} field, while Komi-Zyrian IKDP \cite{Partanen2018} uses \texttt{OrigLang} for Russian words. The KIParla resource includes both code-switching and dialectal variation. We plan to provide information at the word level, through the \texttt{OrigLang} feature that can assume the following values: \texttt{dialect} for dialectal variation, the language iso-639 code in case it exists and \texttt{uncertain} in case the transcriber couldn't assign the word to a specific language.

%but at the moment information is only maintained at the TU level (as opposed to, at the word level), because, specifically in the case of dialectal variation, it is often hard to establish which words can be categorized as standard Italian rather than dialect (e.g., \textit{altro} in the example below pertains to both varieties). Therefore, we introduce a specific metadata tag (\texttt{\# contains\_variation}), that encodes whether the TU contains any variation concerning language varieties.

%\begin{exe}
%\ex 
%\gll \#e tut el rest altro che ades (.) altro che ades\\
%cat.NOM eat.3.SG.PRS sour-cream.ACC sour-cream.ACC sour-cream.ACC sour-cream.ACC\\
%\trans `...'
%\end{exe}

\subsection{Orthography}

Aside from the Jefferson notation, the KIParla project uses standard, written-like \textbf{orthography and spelling}. Generally speaking, Italian orthography does not pose particularly difficult issues. Some (semi-automatic) harmonization will be introduced during revision for words like \textit{okay} (which can be also spelled \textit{okey} for instance) or fillers (see Section \ref{sec:fillers}). No \textbf{capitalization} is performed at transcription step, however, we plan to introduce capitalization in lemmatization for proper nouns. 
An open question concerns the transcription of Italian regional varieties, which often are not standardized, and the transcription of foreign or Italian words pronounced with specific foreign accents that can result in a different orthography.

\subsection{Segmentation}
\textbf{Segmentation} represents the biggest challenge for the KIParla project: boundaries of TUs in fact are highly subjective and were not introduced with the purpose of representing sentence-like units or utterances in any meaningful way. The choice of utterance boundaries clearly influences the possibility to annotate long-distance syntactic relationships holding within the sentence: it is therefore in our opinion crucial to define boundaries in such a way that meaningful relations come to light.
As far as this is concerned, two viable options are being explored. Namely, following the French Raphsodie and Paris Stories corpora \cite{kahane2021annotation}, TUs can be manually aggregated into illocutionary units \cite{cresti1995speech,pietrandrea2014notion} before starting proper syntactic annotation. Alternatively, turns could be taken as base units, furtherly segmenting them only in case of interactions involving one single speaker (e.g., reconrding of lectures).

\textbf{Overlapped speech} is a common phenomenon in KIParla. As we only have alignment at TU level, and sometimes one unit overlaps to multiple other units, we can only provide overlap information at TU level. Therefore, we introduce it in the TU metadata via the \texttt{\# overlap} identifier. Overlapping tokens will then be marked with \texttt{Overlap} feature in the \texttt{MISC} column, albeit without reference to the specific overlapping unit.
For \textbf{co-constructed trees}, we plan on using the \texttt{AttachTo} and \texttt{Rel} features in the \texttt{MISC} field. We however hope that, with more and more spoken language treebanks being introduced among UD resources, a different and shared solution will be discussed for annotating relations that trespass the sentence boundary.

\subsection{Prosody}
Prosody represents the main scope of Jefferson notation. Three symbols are provided to mark descending (\texttt{.}), rising (\texttt{?}) and weakly rising (\texttt{,}) \textbf{intonation patterns}. These are translated into an \texttt{Intonation} feature that can assume values \texttt{Ascending, Descending, Question}. When manually producing morpho-syntactic annotation, these can be used as cues to introduce punctuation.
\textbf{Short pauses} are marked as \texttt{(.)}, consequently the \texttt{PauseAfter=Yes} feature is added in the \texttt{MISC} field on the token preceding the pause.
The Jeffersonian notation also provides information about \textbf{prolonged sounds, volume, pace and the presence of a prosodic link}. Due to the lower agreement among annotators on these traits, we choose not to translate them into explicit features. They remain however present in the \texttt{\# text\_jefferson} metadata field at TU level.

\subsection{Non-Lexical Tokens}
\label{sec:fillers}

\textbf{Fillers or filled pauses} include standard tokens such as \textit{euh} in French, \textit{e} in Norwegian or \textit{ähm} in Turkish-German. The KIParla showcases a number of Italian fillers such as \textit{beh}, \textit{eh}, \textit{ehm} and \textit{mh}.
In existing treebanks, these are either marked as \textit{X} or \textit{INTJ} (we choose the latter) and generally labeled as \textit{discourse} or \textit{discourse:fillers}, attaching to the root of the sentence.

As most other treebanks, the KIParla also includes \textbf{cut-off words}. We align with French Rhapsodie, ParisStories \cite{kahane2021annotation} and Naija NSC \cite{Caron2019} marking them with the $\sim$ symbol at the end of the token in order to avoid any possible overlap with Italian words that contain an hyphen. Speech repairs will be then linked through the \texttt{reparandum} relation to their repair token when the relation to the new word is clear, otherwise they will be marked as \texttt{parataxis:restart} for more complex cases.
%While this is the most logical solution in cases where the relation is clear, speech repairs also often include \textbf{reformulations or replacements} with new words. The Slovenian SST \cite{dobrovoljc2016universal} and NynorskLIA \cite{ovrelid2018lia} treebanks provide a distinction for the two cases, by employing \texttt{parataxis:restart}, \texttt{parataxis:deletion} and \texttt{discourse:filler} relations. We choose this latter approach by keeping \texttt{reparandum} for cases where the relation is clear and either \texttt{parataxis:deletion} or \texttt{parataxis:restart} for more complex cases.
\textbf{Discourse markers} are marked according to their syntactic category (they could be verbal, adverbial, interjections, etc). They are generally labeled as \textit{discourse}, Naija NSC, Slovenian SST and Turkish-German SAGT use \texttt{parataxis:discourse} for distinguishing clausal markers: we follow the same strategy.

%ESEMPIO:
%[sì >sì sì sì<] noi eh mh e::h=mh avremmo cioè l' esame è (a)strutturato ha una traduzione:
%comunque, concetto di idioletto che potrebbe cioè insomma io l' ho connesso soprattutto al concet[to di varie]tà linguistica:: quindi potrebbe anche servirmi per la prima parte.

%\textbf{Incidents} such as laughter and applause are supposedly present in Slovenian SST \cite{dobrovoljc2016universal}. The KIParla transcription showcases a huge variation in how incidents are represented during transcription. They are generally non lexical material but sometimes they interact with the lexical parts. 

%We mark them separately as they do not represent lexical content, but leave a placeholder in the conll in order to mark other information such as pauses etc.

\section{Tentative work plan}

\begin{description}
    \item[December 2024] Selection of a 100K tokens balanced sample of conversations, including a representative subset from each module
    \item[January 2025] Creation of data repository, conversion of current format into \texttt{conll}, data split into \textit{training}, \textit{dev}, \textit{test} sets
    \item[February-March 2025] Automatic Lemmatization and PoS tagging with manual revision, based on KIPoS \cite{Bosco2020} results
    \item[April-June 2025] Manual syntactic annotation
    \item[Summer 2025] First release of the resource
\end{description}

\bibliography{anthology,custom}
\bibliographystyle{acl_natbib}

\end{document}